\documentclass{article}

\usepackage{neurips_2024}

\usepackage[utf8]{inputenc} % allow utf-8 input
\usepackage[T1]{fontenc}    % use 8-bit T1 fonts
\usepackage{hyperref}       % hyperlinks
\usepackage{url}            % simple URL typesetting
\usepackage{booktabs}       % professional-quality tables
\usepackage{amsfonts}       % blackboard math symbols
\usepackage{nicefrac}       % compact symbols for 1/2, etc.
\usepackage{microtype}      % microtypography
\usepackage{xcolor}         % colors
\usepackage{tabulary,multirow,overpic,xcolor}
\usepackage{makecell}
\usepackage{algorithm} % 提供 algorithm 环境
\usepackage{algpseudocode} % 提供伪代码命令
\usepackage{siunitx}  % 提供数值对齐功能
\usepackage{amsmath}
\usepackage{mathtools} % 推荐在导言区加载
\usepackage{multirow}
\usepackage{amssymb}

\title{A Systematic Post-Train
Framework for Video Generation}

\author{%
  Zeyue Xue$^{1,*}$ \And
  Siming Fu$^{2,*}$ \And
  Jie Huang$^{2}$ \And
  Shuai Lu$^{2}$ \And
  Haoran Li$^{2}$ \And
  Yijun Liu$^{3}$  \And
  Yuming Li$^{4}$  \And
  Xiaoxuan He$^{5}$  \And 
  Mengzhao Chen$^{1}$ \And
  Haoyang Huang$^{2}$ \And
  Nan Duan$^{2}$ \And
  Ping Luo$^{1}$ \\\\
  $^1$ The University of Hong Kong~~~~~~ 
  $^2$ JD Explore Academy  \\\\
    $^3$ Tsinghua University~~~~~~ 
  $^4$ Peking University~~~~~~
  $^5$ Zhejiang University \\\\
  * denotes equal contribution.
}

\begin{document}

\maketitle

\begin{abstract}
While large-scale video diffusion models have demonstrated impressive capabilities in generating high-resolution and semantically rich content, a significant gap remains between their pretraining performance and real-world deployment requirements due to critical issues such as prompt sensitivity, temporal inconsistency, and prohibitive inference costs. To bridge this gap, we propose a comprehensive post-training framework that systematically aligns pretrained models with user intentions through four synergistic stages: we first employ Supervised Fine-Tuning (SFT) to transform the base model into a stable instruction-following policy, followed by a Reinforcement Learning from Human Feedback (RLHF) stage that utilizes a novel Group Relative Policy Optimization (GRPO) method tailored for video diffusion to enhance perceptual quality and temporal coherence; subsequently, we integrate Prompt Enhancement via a specialized language model to refine user inputs, and finally address system efficiency through Inference Optimization.  Together, these components provide a systematic approach to improving visual quality, temporal coherence, and instruction following, while preserving the controllability learned during pretraining. The result is a practical blueprint for building scalable post-training pipelines that are stable, adaptable, and effective in real-world deployment.
Extensive experiments demonstrate that this unified pipeline effectively mitigates common artifacts and significantly improves controllability and visual aesthetics while adhering to strict sampling cost constraints.
\end{abstract}

\section{Introduction}
Recent years have seen rapid progress in large-scale diffusion models and diffusion-transformer models \cite{ho2020denoising,esser2024scaling, rombach2022high, lipman2022flow, liu2022flow, gong2025seedream}. These models have advanced from generating short, low-resolution clips to producing longer, higher-resolution videos with more complex motion and richer semantics \cite{gao2025seedance, kong2024hunyuanvideo, team2025kling, wan2025wan}. Despite these improvements, pretrained video generation models still fall short of real-world deployment requirements \cite{huang2024vbench, liu2024evalcrafter}. In practice, they are often sensitive to prompt wording, unstable over long time horizons, prone to local artifacts, such as errors in hands, text, and fast motion, and limited in instruction-following and controllable editing.

This gap between pretraining performance and deployment requirements motivates the need for post-training, which refers to a series of alignment and optimization procedures applied after large-scale likelihood-based training. Unlike pretraining, post-training must operate under strict constraints on sampling cost, evaluation quality, and system efficiency. These challenges are especially severe in video generation, where rollout generation is expensive \cite{xue2025dancegrpo}, and evaluation signals are often noisy \cite{huang2024vbench}.

To address these complexities, we propose a comprehensive post-training paradigm specifically tailored for the video generation lifecycle. Unlike prior approaches that tackle instruction following, visual quality, or inference efficiency in isolation, our framework integrates these objectives into a unified pipeline. By bridging the discrepancy between likelihood-based pretraining and alignment-heavy deployment, we aim to resolve the trade-offs between generation quality, controllability, and system efficiency.

\begin{figure}[t]
  \centering
  % \vspace{-0.5cm}
  \includegraphics[width=1.0\columnwidth]{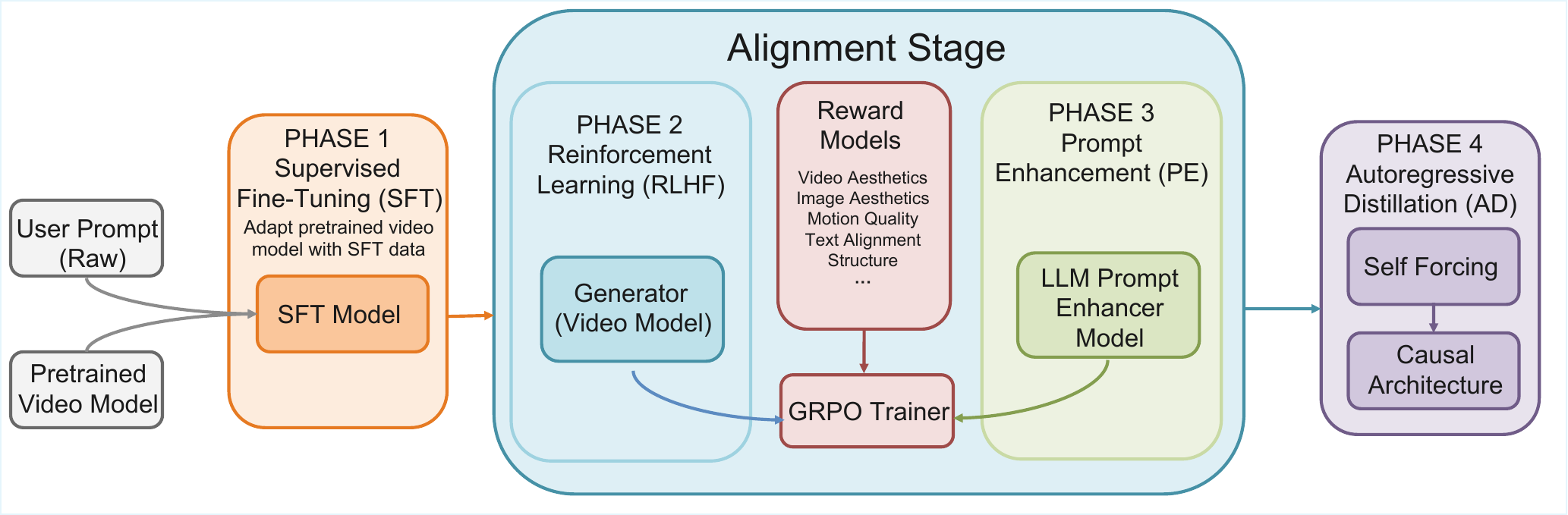}
  \caption{\textbf{Overview of our post-training framework for video generation.} We organize the pipeline into four complementary stages to bridge pretrained models and practical deployment. In \textbf{Phase 1}, supervised fine-tuning (SFT) uses curated data to establish a stable instruction-following baseline. In \textbf{Phase 2}, RLHF via a GRPO-based trainer aligns the generator with multi-dimensional reward signals, improving aesthetics, motion quality, and text alignment. In \textbf{Phase 3}, Prompt Enhancement (PE) optimizes an LLM using the same reward loop to enrich user inputs for better robustness and visual quality. Finally, \textbf{Phase 4} applies autoregressive distillation (AD) with a self-forcing objective to transfer these capabilities into a causal architecture, significantly boosting inference efficiency for real-world deployment.}
  \label{fig:chapter 5 overall}
  % \vspace{-0.5cm}
\end{figure}

As shown in Figure~\ref{fig:chapter 5 overall}, a systematic post-training framework can be organized into \textbf{four} stages:
\begin{itemize}
    \item \textbf{Supervised Fine-Tuning (SFT):} This stage adapts the pretrained model to follow instructions and respond to controllable interfaces. It reduces generation failures and establishes a stable reference policy for later optimization.
    \item \textbf{Reinforcement Learning from Human Feedback (RLHF):} In this stage, we apply a GRPO-based method under a stochastic differential equation formulation to optimize measurable objectives such as perceptual quality and temporal coherence through relative comparisons within prompt groups, without relying on unstable value-function estimation.
    \item \textbf{Prompt Enhancement (PE):} We further use GRPO to train a large language model as a prompt enhancer. The goal is to improve the visual quality of generated outputs while preserving alignment with the original user input.
    \item \textbf{Autoregressive Distillation (AD):} We adopt a self-forcing distillation framework to compress the model into an efficient causal architecture, enabling faster inference while maintaining generation quality.
\end{itemize}

Our key insights are as follows:
\begin{itemize}
    \item \textbf{SFT as the foundation for RLHF:} SFT provides a stable and well-structured policy that makes later reinforcement learning more effective and reliable. SFT also enlarges the exploration for RLHF.
    \item \textbf{Prompt enhancement complements RLHF:} RLHF optimizes the output-side generation policy, while PE refines input-side prompts. Trained with the same rewards—human preference, visual realism, and semantic alignment—PE consistently improves output quality across diverse inputs.
    \item \textbf{Autoregressive distillation enables efficient deployment:} AD transfers the capability of the post-trained generator into a causal architecture, improving inference efficiency while preserving key generation abilities.
\end{itemize}

\section{Related Work}

\subsection{Prompt Enhancement for Visual Generation}
PE for image generation Prompt enhancement (PE) has
become essential for improving text-to-image (T2I) generation quality and alignment \cite{gong2025seedream, gao2025seedream}. While early approaches relied on manual refinement, recent methods leverage LMs
for automated prompt optimization. Promptist \cite{li2024promptist} combines supervised fine-tuning with RL to optimize prompts for aesthetic appeal while preserving user intent. NeuroPrompts \cite{rosenman2024neuroprompts} introduces constrained text decoding for automatic prompt enhancement
with user-controllable styles. OPT2I \cite{manas2024improving}
iteratively refines prompts using LMs to maximize consistency scores. RePrompt \cite{wu2025reprompt} incorporates
chain-of-thought reasoning and reward-guided training for
structured reprompting. PromptRL \cite{wang2026promptrl} proposes a framework that incorporates language models (LMs) as trainable prompt refinement agents directly within the flow-based RL optimization loop. 

\subsection{GRPO for Flow-Matching Models}
Diffusion and flow-matching models 
decompose visual generation into iterative denoising processes, significantly advancing visual synthesis and achieving
state-of-the-art performance in image and video generation. Inspired by the success of reinforcement learning (RL)
in large language models (LLMs), optimization techniques such as PPO \cite{schulman2017proximal} and DPO \cite{rafailov2023direct} have been adapted to diffusion models, facilitating preference alignment and enhancing task-specific outcomes.
In a similar vein, Flow-GRPO \cite{liu2025flow} and DanceGRPO \cite{xue2025dancegrpo} incorporate GRPO-style policy optimization into flow-matching frameworks by reformulating deterministic ODE sampling as stochastic SDE processes, thereby introducing exploratory noise for group-based policy improvement. More recently, MixGRPO \cite{li2025mixgrpo} introduced a hybrid ODE–SDE sampling strategy that enhances training efficiency without compromising generative quality. Concurrently, Flow-CPS \cite{wang2025coefficients} identified a critical limitation in the SDE sampling employed by Flow-GRPO and DanceGRPO, the inconsistent noise coefficients across timesteps, which results in residual noise accumulation and imprecise reward estimation. To mitigate this, Flow-CPS proposes a noise-consistent SDE sampling method that improves reward accuracy and accelerates GRPO convergence. In parallel, TempFlowGRPO \cite{he2025tempflow} and G2RPO \cite{zhou2025text} tackle the issues of reward sparsity and inaccuracy arising from assigning a single global reward to multi-step SDE trajectories. Along the line of addressing sparse/ambiguous supervision over multi-step trajectories, E-GRPO \cite{zhang2026grpo} identifies that only high-entropy steps contribute to effective exploration, and proposes entropy-aware step consolidation with a multi-step group-normalized advantage to improve learning efficiency. BranchGRPO \cite{li2025branchgrpo} reorganizes the rollout process into a branching tree structure, where shared prefixes reduce computational overhead and pruning eliminates low-reward paths and redundant depths. There are some prior arts \cite{zheng2025diffusionnft, xue2025advantage, zhang2026astrolabe} working on forward-process policy optimization.

\subsection{Autoregressive Visual Generation}
To circumvent the limitation of bidirectional diffusion models, autoregressive (AR) approaches enable streaming generation by producing frames sequentially.
While AR models are well-suited for real-time applications, early methods that rely on Teacher Forcing \cite{lamb2016professor} suffer from severe error accumulation during long-video synthesis. 
Recent studies have explored novel training paradigms to address this train-test misalignment. 
Diffusion Forcing \cite{chen2024diffusion} introduces conditioning at arbitrary noise levels, while CausVid \cite{yin2025slow} employs block causal attention and distills bidirectional teacher via distribution matching distillation \cite{yin2024one}. 
More recently, Self-Forcing \cite{huang2025self} and its successors \cite{yang2025longlive, su2026omniforcing, zhu2026causal} establish post-training frameworks that systematically mitigate error accumulation.
Identifying an architectural gap in the initial ODE distillation phase of these frameworks, Causal Forcing reveals that distilling from a bidirectional teacher violates frame-level injectivity. 
Employing an AR teacher for initialization instead, it theoretically bridges this gap to achieve superior real-time generation.

\section{Method}
\subsection{SFT as the Foundation for RLHF}
In our framework, supervised fine-tuning (SFT) is not intended to fully solve alignment or optimize subjective quality. Instead, its main role is to establish a stable and well-structured reference policy that supports all subsequent post-training stages. This is a deliberate design choice, as SFT addresses the critical “low-hanging fruit” of model behavior, transforming a potentially erratic and unpredictable policy into one that is coherent and structurally sound. During the SFT phase, we systematically target and eliminate the most severe and frequent failures, such as refusal cascades, incoherent reasoning, and unsafe outputs, thereby creating a reliable baseline. This baseline is essential for the success of later stages like RLHF, as it provides a stable starting point that prevents the model from diverging into degenerate behaviors during further optimization. By ensuring the model first learns to follow instructions and maintain basic safety, SFT enables more efficient and effective refinement of nuanced alignment and subjective quality in subsequent phases, ultimately leading to a more robust and capable model.

\subsection{GRPO for Flow-Matching Models and Prompt Enhancer}
\subsubsection{GRPO for Flow-Matching Models}

Following DanceGRPO \cite{xue2025dancegrpo}, we formulate the sampling process of flow-matching models under stochastic dynamics as a Markov decision process (MDP), defined by
\(
(\mathcal{S}, \mathcal{A}, \rho_0, P, \mathcal{R})
\).
Under this formulation, the policy induces a trajectory over the discrete sampling process:
\[
\Gamma = (\mathbf{s}_0, \mathbf{a}_0, \mathbf{s}_1, \mathbf{a}_1, \ldots, \mathbf{s}_T, \mathbf{a}_T).
\]
We consider a sparse-reward setting in which supervision is provided only at the terminal step. Specifically, the reward function is defined as:
\[
\mathcal{R}(\mathbf{s}_i, \mathbf{a}_i) \triangleq
\begin{cases}
R(\mathbf{x}_T, c), & i = T,\\
0, & \text{otherwise},
\end{cases}
\]
where \(R(\mathbf{x}_T, c)\) denotes the reward assigned by the reward model to the final generated sample \(\mathbf{x}_T\) conditioned on the prompt \(c\).

For deterministic reverse-time sampling, the probability flow ODE is given by:
\begin{equation}
\label{eq:prob_ode}
\frac{\mathrm{d}\mathbf{x}_t}{\mathrm{d}t}
=
f(\mathbf{x}_t,t)
-
\frac{1}{2} g^2(t)\nabla_{\mathbf{x}_t}\log q_t(\mathbf{x}_t),
\end{equation}
where \(q_t(\mathbf{x}_t)\) denotes the marginal distribution at time \(t\), and \(\nabla_{\mathbf{x}_t}\log q_t(\mathbf{x}_t)\) is the corresponding score function.

According to the Fokker--Planck equation, Eq.~\eqref{eq:prob_ode} admits an equivalent reverse-time SDE that preserves the same marginal distribution at each time \(t\):
\begin{equation}
\label{eq:prob_sde}
\mathrm{d}\mathbf{x}_t
=
\left[
f(\mathbf{x}_t, t)
-
g^2(t)\nabla_{\mathbf{x}_t}\log q_t(\mathbf{x}_t)
\right]\mathrm{d}t
+
g(t)\mathrm{d}\mathbf{w}_t,
\end{equation}
where \(\mathbf{w}_t\) denotes a standard Wiener process.

MixGRPO \cite{li2025mixgrpo} adopts a hybrid sampling strategy that combines ODE and SDE updates. Formally, the mixed sampling process is defined as:
\begin{small}
\begin{equation}
\label{eq:mix_sampling}
\mathrm{d}\mathbf{x}_t =
\begin{cases}
\left[f(\mathbf{x}_t, t) - g^2(t)\mathbf{s}_t(\mathbf{x}_t)\right]\mathrm{d}t + g(t)\mathrm{d}\mathbf{w}_t,
& \text{if } t \in S, \\[4pt]
\left[f(\mathbf{x}_t, t) - \frac{1}{2}g^2(t)\mathbf{s}_t(\mathbf{x}_t)\right]\mathrm{d}t,
& \text{otherwise},
\end{cases}
\end{equation}
\end{small}
where
\(
\mathbf{s}_t(\mathbf{x}_t) \triangleq \nabla_{\mathbf{x}_t}\log q_t(\mathbf{x}_t)
\)
denotes the score function, and \(S\) is the subset of time steps at which stochastic updates are applied.

However, when applied to video generation, MixGRPO tends to suffer from reward collapse when the stochastic subset is small. To reduce the substantial computational cost of video generation, and motivated by Flash-GRPO, we adopt isotemporal grouping, in which each prompt is assigned a distinct timestep \(t_i\). During denoising, each prompt group performs a single ODE-to-SDE transition at its assigned timestep \(t_i\). The selected timestep uses SDE sampling to enable exploration and gradient computation, whereas all other timesteps use deterministic ODE updates to produce higher-quality generations and more reliable reward signals. We further adopt Temporal Gradient Rectification to explicitly normalize the time-dependent scaling factor:
\[
\lambda(t) = \frac{\sqrt{\Delta t}}{\sigma_t} + \frac{\sigma_t \sqrt{\Delta t}(1 - t)}{2t}.
\]

Based on the mixed sampling formulation above, we optimize the model using a GRPO-style objective. Given a prompt \(c \sim \mathcal{C}\), we first sample a group of \(N\) trajectories from the reference policy \(\pi_{\theta_{\text{old}}}(\cdot \mid c)\), and then optimize:
\begin{small}
\begin{equation}
\label{eq:mixgrpo_objective}
\begin{aligned}
\mathcal{J}_{\text{Flash-GRPO}}(\theta)
=
\mathbb{E}_{c\sim\mathcal{C},~\{\mathbf{x}_T^i\}_{i=1}^{N}\sim\pi_{\theta_{\text{old}}}(\cdot|c)}
\left[
\frac{1}{N}\sum_{i=1}^{N}
\min\!\left(
\frac{r_{t}^{i}(\theta)}{\lambda(t)}A^i,\,
\operatorname{clip}\!\bigg(\frac{r_{t}^{i}(\theta)}{\lambda(t)}, 1-\varepsilon, 1+\varepsilon\bigg)A^i
\right)
\right],
\end{aligned}
\end{equation}
\end{small}
where \(\varepsilon\) is the clipping coefficient, \(r_t^i(\theta)\) is the policy ratio, and \(A^i\) denotes the group-normalized advantage. We compute this policy loss at timestep \(t_i\) for each rollout. More specifically, these quantities are defined as:
\begin{equation}
\label{eq:mixgrpo_terms}
\begin{aligned}
r_t^i(\theta)
&=
\frac{
q_{\theta}(\mathbf{x}_{t+\Delta t}\mid \mathbf{x}_t, c)
}{
q_{\theta_{\text{old}}}(\mathbf{x}_{t+\Delta t}\mid \mathbf{x}_t, c)
}, \\
A^i
&=
\frac{
R(\mathbf{x}_T^i, c)
-
\operatorname{mean}\!\left(\{R(\mathbf{x}_T^i, c)\}_{i=1}^{N}\right)
}{
\operatorname{std}\!\left(\{R(\mathbf{x}_T^i, c)\}_{i=1}^{N}\right)
}.
\end{aligned}
\end{equation}

The objective in Eq.~\eqref{eq:mixgrpo_objective} encourages reward improvement through terminal feedback while constraining policy updates via clipping. In this way, the proposed framework achieves a favorable balance between optimization stability and reward-driven exploration for flow-matching models.

Building on DanceGRPO \cite{xue2025dancegrpo}, we omit the KL regularization term and adopt a strategy analogous to HPSv3 \cite{ma2025hpsv3}. Specifically, using these methods \cite{xu2023imagereward, kirstain2023pick, liu2025improving, wu2023human, wu2023human1, he2024videoscore, xu2026visionreward, wang2025unified, wu2025rewarddance} as references, we adopt a two-stage training framework. In Stage 1, we employ data-aware orthogonal gradient projection to integrate diverse aesthetic preferences derived from HPDv3++ while preserving the original human preference knowledge encoded in HPSv3. In Stage 2, we further leverage unlabeled data generated by models with varying capability levels and from different RL iterations. We use these four reward models:
\begin{itemize}
    \item \textbf{Video Aesthetics:} Evaluates the overall visual quality of generated videos, including lighting, composition, color harmony, temporal consistency, and cinematic appearance across frames.
    
    \item \textbf{Image Aesthetics:} Measures frame-level perceptual quality and aesthetic appeal, encouraging sharp details, pleasing structure, and high-quality visual rendering in individual key frames.
    
    \item \textbf{Motion Quality:} Assesses the realism, smoothness, and coherence of motion dynamics, reducing artifacts such as jitter, discontinuous movement, or temporally inconsistent object transitions.
    
    \item \textbf{Text-Video Alignment:} Evaluates the semantic consistency between the input prompt and the generated video, ensuring that the generated content faithfully reflects the described objects, actions, scenes, and overall intent of the prompt.
\end{itemize}

Integrating these reward models into a unified RL framework is highly nontrivial. Unlike single-reward optimization, our setting requires jointly handling multiple reward signals with different granularities, scales, and optimization tendencies. For example, emphasizing text-video alignment may improve semantic fidelity but can sometimes hurt visual naturalness, while overly prioritizing motion quality or video aesthetics may lead to visually pleasing yet semantically weaker generations. Therefore, a key challenge of the system lies in balancing the relative contributions of the four reward models during training.

To address this issue, we carefully design the reward aggregation strategy and tune the weighting coefficients among different reward components, so that the optimization process remains stable while avoiding domination by any single objective. In practice, we find that properly balancing these reward models is crucial for achieving high-quality video generation. Our final system is designed to trade off semantic accuracy, motion consistency, frame-level fidelity, and overall video aesthetics, ultimately yielding the best visual quality as the primary objective.

\subsubsection{GRPO for Prompt Enhancer}
To achieve this, we adopt the RePrompt \cite{wu2025reprompt} paradigm, treating the prompt optimization process as a reinforcement learning problem. We design a composite reward mechanism comprising three distinct objectives to guide the policy:

\begin{itemize}
    \item \textbf{Text-Video Alignment:} Ensures semantic consistency between the generated content and the input prompt.
    \item \textbf{Video Aesthetics:} Evaluates visual quality, including lighting, composition, and temporal coherence.
    \item \textbf{Structure Reward:} Enforces structural constraints (e.g., format compliance, length) to ensure the prompt is valid and executable.
\end{itemize}

The optimization objective is formulated using Group Relative Policy Optimization (GRPO), which eliminates the need for a value network by leveraging group-based advantage estimation. The objective function is defined as:

\begin{footnotesize}
\begin{equation}
\begin{aligned}
\mathcal{J}_{\mathrm{GRPO}}&(\theta) \\
= &\mathbb{E}_{P,\{y_i\}_{i=1}^G \sim\pi_{\theta_{\mathrm{old}}}} \Bigg[\,
\frac{1}{G}\sum_{i=1}^G 
\min\!\Big(r_i\,A_i,\;\mathrm{clip}(r_{i},1-\varepsilon,1+\varepsilon)\,A_i\Big) \nonumber
- \beta_{\mathrm{KL}}\;D_{\mathrm{KL}}\big(\pi_\theta(y\mid P)\,\|\,\pi_{\mathrm{ref}}(y\mid P)\big)\Bigg],
\end{aligned}
\label{eq:grpo}
\end{equation}
\end{footnotesize}

where $P$ denotes the input prompt context and $y$ is the output, and $\{y_i\}_{i=1}^G$ represents a group of $G$ outputs sampled from the old policy $\pi_{\theta_{\mathrm{old}}}$. The term $r_i = \frac{\pi_\theta(y_i\mid P)}{\pi_{\theta_{\mathrm{old}}}(y_i\mid P)}$ represents the probability ratio. Crucially, the advantage $A_i$ is computed based on the normalized rewards within the group, encouraging the model to prioritize high-performing prompts relative to their peers. $\varepsilon$ and $\beta_{\mathrm{KL}}$ serve as clipping and KL-penalty coefficients to stabilize training.

By freezing the generative backbone and exclusively optimizing the policy $\pi_\theta$, RePrompt functions as a universal framework. It can be seamlessly applied to any off-the-shelf generative model (T2I or T2V), learning specific reasoning patterns and prompt strategies without the computational burden of retraining the underlying image or video generator.

\subsection{Autoregressive Distillation}
An autoregressive video diffusion model is a hybrid generative framework that integrates autoregressive chain-rule decomposition with denoising diffusion for video generation.
Formally, given a sequence of $N$ video frames $x^{1:N} = (x^1, x^2, \dots, x^N)$, their joint distribution can be factorized using the chain rule: $p(x^{1:N}) = \prod_{i=1}^{N} p(x^i \mid x^{<i}).$ Autoregressive video diffusion models are trained by distilling a pretrained bidirectional model. With this model, we use a Distribution Matching Distillation (DMD) loss:
\begin{equation}
\label{eq:dmd}
\nabla_{\theta} \mathcal{L}_{\text{DMD}} 
 \approx - \mathbb{E}_{t} \left( \int \left( s_{\text{data}} \left( \Psi \left( G_{\theta}(\epsilon), t \right), t \right) - s_{\text{gen}} \left( \Psi \left( G_{\theta}(\epsilon), t \right), t \right) \right) \frac{d G_{\theta}(\epsilon)}{d \theta} \, d\epsilon \right), 
\end{equation}
where $\Psi$ represents the forward diffusion process, $\epsilon$ is random Gaussian noise, $G_{\theta}$ is the generator parameterized by $\theta$, and $s_{\text{data}}$ and $s_{\text{gen}}$ represent the score functions trained on the data and generator's output distribution, respectively. 

Our training consists of three stages:
\begin{itemize}
\item \textbf{Training with DMD.} We first employ DMD to distill the original pretrained model into a bidirectional student model that requires only a few denoising steps. While preserving the original global-attention receptive field, this stage equips the model with strong few-step denoising capability, thereby providing a high-quality and easily regressible teacher trajectory for the subsequent migration to a causal architecture.

\item \textbf{Causal ODE Regression.} Directly training the causal student model with the DMD loss can be unstable because of architectural discrepancies. To address this issue, we introduce an efficient initialization strategy to stabilize training and equip the model with block-causal masks. This stage aims to facilitate causal adaptation by training the model to make effective denoising predictions based solely on causal history.

\item \textbf{Self-Forcing Distillation.} We adopt a Self-Forcing distillation paradigm, in which each frame is generated conditioned on previously self-generated outputs through autoregressive rollout with key-value (KV) cache during training. This strategy enables supervision via a DMD loss at the video level, thereby directly evaluating the quality of the entire generated sequence.
\end{itemize}
For models with audio-visual generation, we follow Omniforcing \cite{su2026omniforcing}, by equipping the model with asymmetric block-causal alignment and an audio sink token.

\section{Experiments}
\subsection{Setup}
\textbf{Experimental Settings.} We utilize an internal video generation model. We have trained several distinct types of reward models. Furthermore, we curated a specific prompt set for RLHF training, DMD, ODE regression, and self-forcing training.

\textbf{Dataset.} 
First, we constructed a high-quality text-video dataset for SFT. Subsequently, we curated the prompt set as described in the experimental settings.

\textbf{Reward Models.} We follow the HPSv3~\cite{ma2025hpsv3} training paradigm, using Qwen3.5~\cite{bai2025qwen3} as the backbone to extract features from both images and text. These features are processed through a Multilayer Perceptron (MLP) to produce the final output. 
In our approach, for a given pair of training images $(x_1, x_2)$ with their corresponding text prompt $c$ and human preference annotation $(y_1, y_2)$, we derive the reward scores using the following equations:
\begin{equation}
    r_1 = f_\phi(\mathcal{E}_{\theta}(x_1,c)),
    r_2 = f_\phi(\mathcal{E}_{\theta}(x_2,c)).
\end{equation}
Here, $\mathcal{E}_{\theta}$ denotes the vision-language model, and $f_\phi$ refers to the MLP. Moreover, we adopt an uncertainty-aware ranking loss. We collected a dataset covering video aesthetics, text-video alignment, image aesthetics, and text-image alignment, resulting in four distinct reward metrics.

\textbf{Evaluation Protocol.} Following recent large-scale video generation reports, we adopt a Good–Same–Bad (GSB) comparison protocol. GSB is well-suited for video evaluation, as it explicitly allows annotators to express indifference when differences are subtle, thereby reducing forced or noisy
decisions in marginal cases.

\textbf{Evaluation Aspects}: We evaluate three complementary aspects that jointly characterize video generation quality:
\begin{itemize}
    \item Visual quality: overall appearance, sharpness, and absence of visual artifacts;
    \item Motion quality: temporal coherence, smoothness, and plausibility of motion patterns;
    \item  Text alignment: consistency between the generated video and the input prompt semantics;
\end{itemize}
In this report, we ask the human artist to give an overall comparsion for the results.

\begin{figure}[htbp]
  \centering
  % \vspace{-0.5cm}
  \includegraphics[width=1.0\columnwidth]{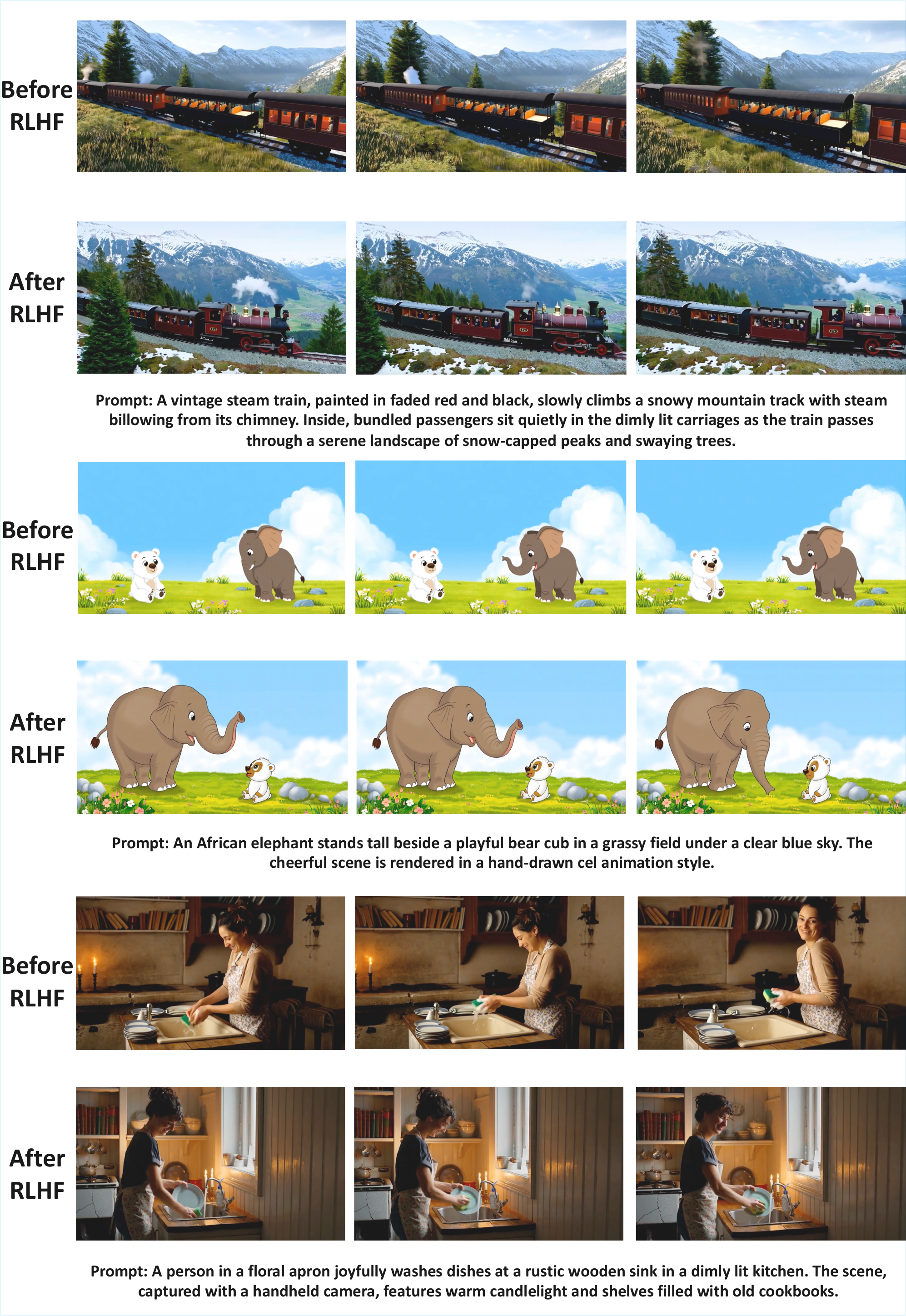}
  \caption{The visualization of RLHF on Wan-2.1.}
  \label{fig:wan-rlhf}
  % \vspace{-0.5cm}
\end{figure}

\subsection{Results}
For our internal model, our RLHF method achieves a substantial ~31\% improvement in the overall GSB metric. When breaking down the performance across specific dimensions, the gains are most pronounced in visual quality and motion quality, both of which exhibit massive enhancements. In contrast, the improvement in text alignment is relatively modest. We attribute this discrepancy to the limited accuracy of the current text alignment reward model, which restricts the optimization potential for semantic correctness. Furthermore, the integration of the prompt enhancer yields an additional ~20\% improvement in overall GSB. This strong preference is similarly driven by significant improvements in visual and motion quality, while preserving text alignment. Together, these results demonstrate that our framework substantially improves aesthetic appearance and temporal dynamics without compromising the established baseline of semantic alignment.
The example visualization of RLHF could be found on Figure~\ref{fig:wan-rlhf}.

\section{Conclusion}
In this paper, we proposed a comprehensive and unified post-training framework to bridge the critical gap between the pretraining capabilities of large-scale video diffusion models and the rigorous demands of real-world deployment. By systematically integrating four synergistic stages, Supervised Fine-Tuning (SFT), Reinforcement Learning from Human Feedback (RLHF) via a novel Group Relative Policy Optimization (GRPO) method, Prompt Enhancement (PE), and Autoregressive Distillation (AD), our pipeline effectively mitigates common local artifacts, temporal inconsistencies, and high inference costs.

Extensive human evaluations utilizing the Good–Same–Bad (GSB) protocol demonstrate the profound efficacy of our approach. Our RLHF stage achieved a substantial ~31\% improvement in the overall GSB metric, driven by massive enhancements in visual quality and motion coherence. Furthermore, the integration of our specialized prompt enhancer yielded an additional ~20\% overall GSB improvement, elevating perceptual aesthetics and temporal dynamics while strictly preserving the baseline semantic alignment.

While the framework significantly enhances generation quality and controllability, the relatively modest improvements in text alignment highlight the limitations of current text-video reward models. Future work will focus on developing more robust and accurate text alignment reward models to fully unlock the optimization potential for semantic correctness. Ultimately, this work provides a scalable, adaptable, and highly practical blueprint for building deployable video generation pipelines that successfully balance visual excellence, temporal consistency, and system efficiency.

\section{Broader Impact}
The proposed post-training framework significantly bridges the gap between foundational video diffusion models and practical deployment, unlocking transformative applications across e-commerce, digital marketing, entertainment, and the broader creative industries through scalable, high-fidelity, and computationally efficient video synthesis. By making advanced video generation systems more reliable and adaptable to real-world production demands, this framework not only improves output quality but also lowers the barrier for integrating generative video technologies into commercial content pipelines, personalized advertising, and interactive media experiences. In this sense, it represents an important step toward turning foundation-level video models into deployable and economically valuable infrastructure.

Furthermore, this strict optimization for continuous temporal dynamics, fine-grained controllability, and complex instruction alignment forces the model to internalize more accurate physical laws, stronger object permanence, and more stable representations of causal interactions over time. Rather than merely improving surface-level perceptual quality, the framework contributes to deeper generative competence by encouraging the model to preserve structural consistency and event logic throughout evolving visual sequences. This substantially advances the foundational capabilities required for robust Video World Models, especially in settings where long-range temporal reasoning and environment persistence are essential.

\bibliographystyle{unsrt}
\bibliography{main}

%%%%%%%%%%%%%%%%%%%%%%%%%%%%%%%%%%%%%%%%%%%%%%%%%%%%%%%%%%%%

%%%%%%%%%%%%%%%%%%%%%%%%%%%%%%%%%%%%%%%%%%%%%%%%%%%%%%%%%%%%

\end{document}